\documentclass[10pt,twocolumn,letterpaper]{article}

\usepackage{iccv}
\usepackage{times}
\usepackage{epsfig}
\usepackage{graphicx}
\usepackage{amsmath}
\usepackage{amssymb}

\usepackage{booktabs}
\usepackage{bbm}
\newcommand{\thename}{HAIS}
\def\eg{\emph{e.g}\onedot} 
\def\ie{\emph{i.e}\onedot} 

\def\etal{\emph{et al}\onedot}
\usepackage{makecell}
\usepackage[ruled,linesnumbered]{algorithm2e}

\usepackage{subfigure}
\usepackage[breaklinks=true, bookmarks=false, colorlinks=true]{hyperref}

\iccvfinalcopy % *** Uncomment this line for the final submission

\def\iccvPaperID{4133} % *** Enter the ICCV Paper ID here

\ificcvfinal\fi

\begin{document}

%%%%%%%%% TITLE
\title{Hierarchical Aggregation for 3D Instance Segmentation}

\author{
{Shaoyu Chen}\textsuperscript{1} \quad {Jiemin Fang}\textsuperscript{2,1} \quad {Qian Zhang}\textsuperscript{3} \quad {Wenyu Liu}\textsuperscript{1} \quad {Xinggang Wang}\textsuperscript{1}$^\dag$\\ 
\textsuperscript{1} School of EIC, Huazhong University of Science \& Technology \\
 \textsuperscript{2} Institute of AI, Huazhong University of Science \& Technology\quad \textsuperscript{3} Horizon Robotics\\

  {\tt\small \{shaoyuchen,jaminfong,liuwy,xgwang\}@hust.edu.cn\quad \{qian01.zhang\}@horizon.ai}
}

% \author{%
%  \textbf{Zhongqi Yue}\textsuperscript{1,3}\thanks{Equal contribution}, \quad \textbf{Tan Wang}\textsuperscript{1}\footnotemark[1], \quad \textbf{Hanwang Zhang}\textsuperscript{1}, \quad \textbf{Qianru Sun}\textsuperscript{2}, \quad \textbf{Xian-Sheng Hua}\textsuperscript{3}\\
% \small \textsuperscript{1}Nanyang Technological University,\quad \textsuperscript{2}Singapore Management University, \small\textsuperscript{3}Damo Academy, Alibaba Group\\
% \tt\small yuez0003@ntu.edu.sg,\quad TAN317@e.ntu.edu.sg,\quad hanwangzhang@ntu.edu.sg,\\
% \tt\small \quad qianrusun@smu.edu.sg, \quad xiansheng.hxs@alibaba-inc.com\\}

\maketitle

%%%%%%%%% ABSTRACT
\begin{abstract}
Instance segmentation on point clouds is a fundamental task in 3D scene perception. In this work, we propose a concise clustering-based framework named \thename, which makes full use of spatial relation of points and point sets. Considering clustering-based methods may result in over-segmentation or under-segmentation, we introduce the hierarchical aggregation to progressively generate instance proposals, \ie, point aggregation for preliminarily clustering points to sets and set aggregation for generating complete instances from sets.
Once the complete 3D instances are obtained, a sub-network of intra-instance prediction is adopted for noisy points filtering and mask quality scoring. \thename\ is fast (only $410ms$ per frame) and does not require non-maximum suppression. It ranks 1st on the ScanNet v2 benchmark
$\footnote{\url{http://kaldir.vc.in.tum.de/scannet_benchmark/semantic_instance_3d}}$, 
achieving the highest $69.9\%$ $AP_{50}$ and surpassing previous state-of-the-art (SOTA) methods by a large margin. Besides, the SOTA results on the S3DIS dataset validate the good generalization ability.
Code will be available at \url{https://github.com/hustvl/HAIS}.
\end{abstract}
%%%%%%%%% BODY TEXT

\begin{figure}[thbp]
    \centering
    \includegraphics[width=\linewidth]{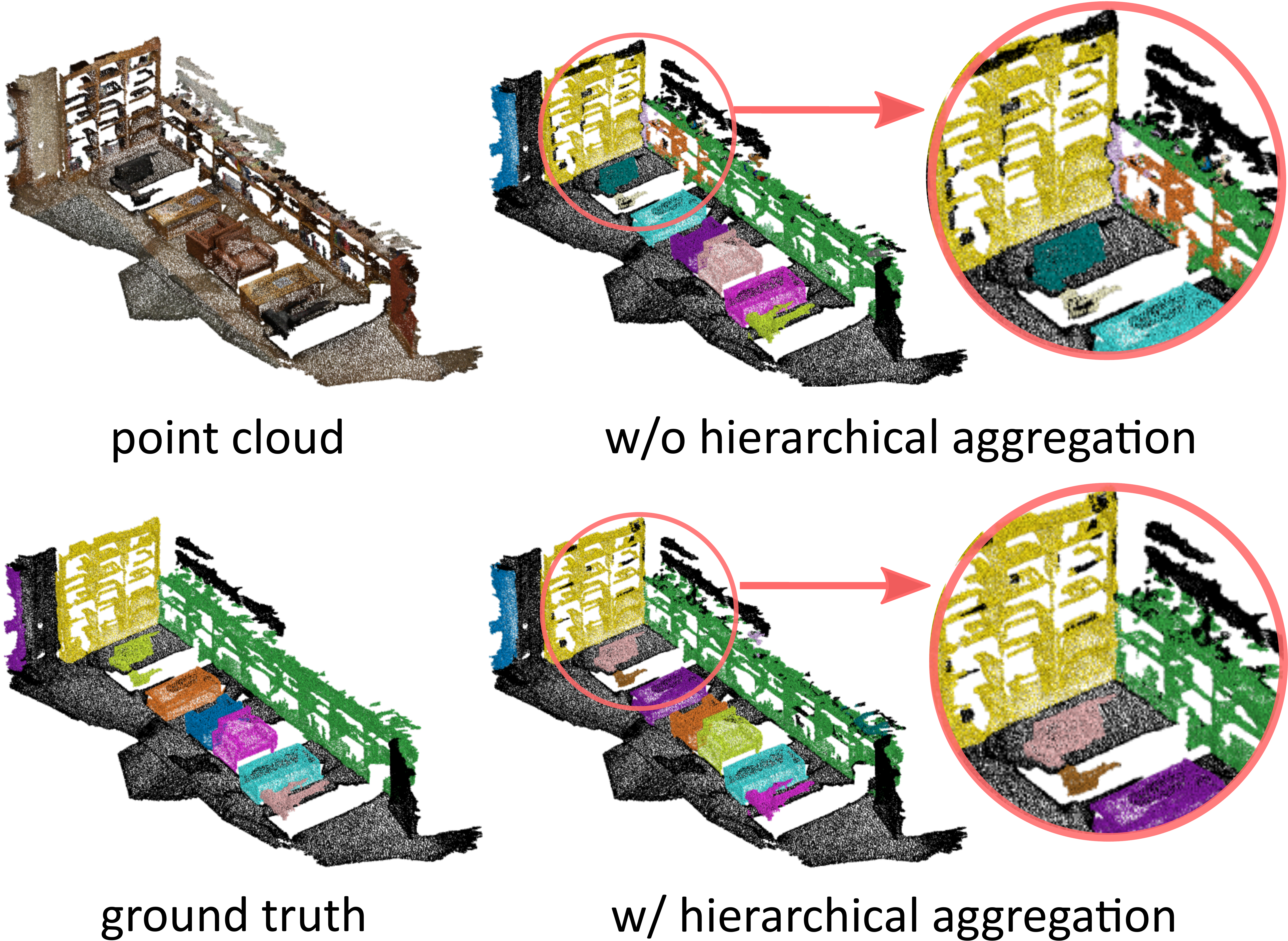}
    \caption{An input point cloud, ground truth instance masks and 3D instance prediction results without \& with hierarchical aggregation. As shown in the key region circled with red, for objects with large sizes and fragmentary point clouds,
    the predictions are easy to be over-segmented. The proposed hierarchical aggregation combines incomplete instances with fragments to form complete instance predictions. }
    \label{fig:introduction} 
\end{figure}

\let\thefootnote\relax\footnotetext{$^\dag$Xinggang Wang is the corresponding author.}

\section{Introduction}
With the rapid development and popularization of commodity 3D sensors (Kinect, RealSense, Velodyne laser scanner, \etc), 3D scene understanding has become a hot research topic in the field of computer vision. Instance segmentation on point cloud, as the basic perception task of 3D scene understanding, is the technical foundation of a wide range of real-life applications, \eg, robotics, augmented/virtual reality, and autonomous driving.
 
Instance segmentation on 2D images has been exhaustively studied in the past few years~\cite{ISFCN,FCIS,MaskRCNN,PANet,MSRCNN,HTC,BMaskRCNN, QueryInst}. Top-down methods dominate 2D instance segmentation. They first generate instance-level proposals and then predict the mask for each proposal.
Though existing 2D instance segmentation methods can be directly extended to 3D scenes, most existing 3D methods adopt a totally different bottom-up pipeline~\cite{ASIS,SGPN,MTML,OccuSeg,PointGroup}, which generates instances through clustering.

However, directly clustering a point cloud into multiple instances is very difficult for the following reasons: $(1)$ A point cloud usually contains a large number of points; $(2)$ The number of instances in a point cloud has large variations for different 3D scenes; $(3)$ The sizes of instances vary significantly; $(4)$ Each point has a very weak feature, \ie, 3D coordinate and color. The semantic gap between point and instance identity is huge.
Thus, over-segmentation or under-segmentation are common problems and are prone to exist.

We propose a hierarchical aggregation scheme in bottom-up 3D instance segmentation networks to cope with these problems.
We first aggregate points to sets with low bandwidth to avoid over-segmentation and then set aggregation with dynamic bandwidth is adopted to form complete instances. 
Set aggregation may absorb noisy point sets into predictions, making the aggregated instances over-complete. Thus, we design a sub-network for outlier filtering and mask quality scoring. Based on the hierarchical aggregation and the sub-network for intra-instance prediction, we propose a novel bottom-up framework, named \thename. 
\thename\ achieves the state-of-the-art (SOTA) performance on both the ScanNet v2 benchmark~\cite{ScanNet} and the S3DIS~\cite{S3DIS} dataset. 
Beyond this, \thename\ is efficient, only requiring a concise single-forward inference without any post-processing steps. Compared with all the existing methods, \thename\ takes the lowest inference latency.

Our contributions can be summarized as follows.
\begin{itemize}
\item We propose a novel bottom-up framework with the hierarchical aggregation for instance segmentation on 3D point cloud. The hierarchical aggregation strategy makes up the defects of bottom-up clustering. Besides, an intra-instance prediction network is designed for generating more fine-grained instance predictions.
\item Our method ranks 1st on the leaderboard of ScanNet v2~\cite{ScanNet}. \thename\ also achieves the state-of-the-art result on S3DIS~\cite{S3DIS}. We significantly promote the performances on various challenging datasets and demonstrate the generalization of the proposed methods.
\item  Our method achieves the highest efficiency among all existing methods. \thename\ keeps a concise single-forward inference pipeline without any post-processing steps. The average per-frame inference time on ScanNet v2 is only $410$ ms, much faster than other methods. 
\end{itemize}

\section{Related Works}
\paragraph{Deep Learning on Point Clouds}
Extracting features from point clouds is the foundation of 3D scene understanding. Deep learning equipped methods mainly include point-based ones and voxel-based ones. Point-based methods, \eg, PointNet~\cite{PointNet} and PointNet++~\cite{PointNet++}, directly operate on unstructured sets of points. Voxel-based methods~\cite{Submanifold, OctNet, SEGCloud, VoxNet} transform the unordered and unstructured point sets to ordered and structured volumetric grids, and then perform 3D sparse convolutions on the grids. We adopt voxel-based methods for more efficient feature extraction.  

\paragraph{Proposal-based Instance Segmentation}
Proposal-based approaches directly generate object proposals and predict masks inside each proposal.
In the 2D domain, proposal-based methods employ 2D object detectors~\cite{FastRCNN,FasterRCNN,RFCN,FPN} to generate region proposals and then predict masks inside each proposal. Mask R-CNN~\cite{MaskRCNN} extends Faster R-CNN~\cite{FasterRCNN} by adding a mask prediction. 
% HTC~\cite{HTC} interweaves box and mask branches in a multi-stage cascade manner. 
EmbedMask~\cite{EmbedMask} introduces proposal embedding and pixel embedding so that pixels are assigned to instance proposals according to their embedding similarity.
In the 3D domain, GSPN~\cite{GSPN} proposes a generative shape proposal network for 3D object proposals following an analysis-by-synthesis strategy. 3D-SIS~\cite{3DSIS} takes both 3D geometry and 2D color images as input and combines 2D and 3D features through the back projection for a better prediction. 
3D-BoNet~\cite{BoNet} regresses a fixed set of bounding boxes and designs a novel association layer to match predicted boxes and ground truth boxes.
3D-MPA~\cite{3D-MPA} predicts centers of instances and employs a graph convolutional network to refine proposal features.
GICN~\cite{GICN} approximates the distributions of instance centers as Gaussian center heatmaps and uses a center selection mechanism for choosing candidates.

\begin{figure*}[htb]
\centering
\includegraphics[width=\linewidth]{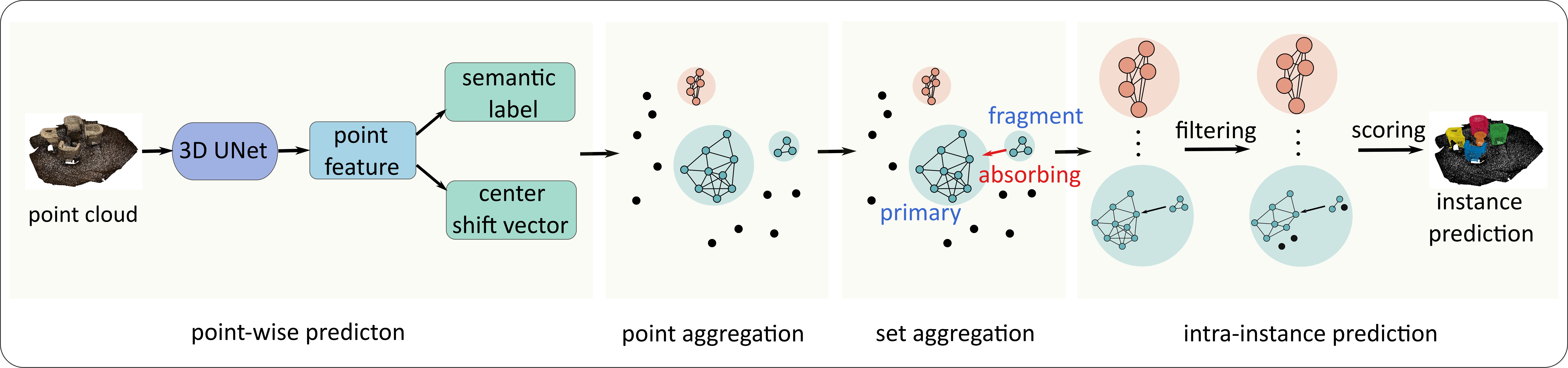}
\caption{The framework of \thename. For the input point cloud, our method first employs 3D UNet-like structure with submanifold sparse convolution~\cite{UNet,Submanifold} for point-wise feature learning. Then, we use the spatial constraint of points to perform point aggregation with fixed bandwidth. Based on point aggregation results, set aggregation with dynamic bandwidth is performed to form instance proposals. The intra-instance prediction is designed for outlier filtering and mask quality scoring.}
\label{fig:framework}
\end{figure*}

\paragraph{Clustering-based Instance Segmentation}
Clustering-based approaches first predict point-wise labels and then use clustering methods to generate instance predictions.
In the 2D domain, metric learning is widely used to group pixels.
Fathi \etal~\cite{FathiWRWSGM17} compute likelihoods of pixels and group similar pixels together within an embedding space.
Bai and Urtasun~\cite{BaiU17}  adopt energy maps to distinguish among individual instances.
Kong and Fowlkes~\cite{KongF18a} assign all pixels to a spherical embedding for clustering. 
Neven \etal~\cite{NevenBPG19} introduce a learnable clustering bandwidth instead of learning the embedding using hand-crafted cost functions.
Bert \etal~\cite{Bert2017} propose a discriminative loss function which encourages the network to map each pixel to a point in feature space so that pixels belonging to the same instance lie close together while different instances are separated by a wide margin. 
In the 3D domain, SGPN~\cite{SGPN} proposes to learn a similarity matrix for all point pairs and merges similar points to generate instances.
JSIS3D~\cite{JSIS3D} adopts multi-value conditional random fields to form instance predictions.
MTML~\cite{MTML} introduces a multi-task learning strategy for grouping points.
OccuSeg~\cite{OccuSeg} adopts learnt occupancy signals to guide clustering.
PointGroup~\cite{PointGroup} proposes to cluster points based on dual coordinate sets and designs ScoreNet to predict scores for instances.

Our \thename\ follows the clustering-based paradigm but differs from existing clustering-based methods in two terms.
First, most clustering-based methods require complicated and time-consuming clustering procedures, but our \thename\ adopts a much more concise pipeline and keeps high efficiency.
Second, previous methods usually group points according to point-level embeddings, without the instance-level correction.
Our \thename\ introduces the set aggregation and intra-instance prediction to refine the instance at the object level.

\section{Method}
The overall architecture of \thename\ is depicted in Fig.~\ref{fig:framework}, which consists of four main parts. The point-wise prediction network (Sec.~\ref{sec:point-wise}) extracts features from point clouds and predicts point-wise semantic labels and center shift vectors. The point aggregation module (Sec.~\ref{sec:point-aggr}) forms preliminary instance predictions based on point-wise prediction results. The set aggregation module (Sec.~\ref{sec:set-aggr}) expands incomplete instances to cover missing parts, while the intra-instance prediction network (Sec.~\ref{sec:intra}) smooths instances to filter out outliers.

\subsection{Point-wise Prediction Network}
\label{sec:point-wise}
The point-wise prediction network takes the point cloud $P \in \mathbb{R}^{N \times K}$ as input, where $N$ is the number of points and $K$ is the number of channels. $K$ is normally set as 6 for colors $r$, $g$, $b$ and locations $x$, $y$, $z$.
The submanifold sparse convolution~\cite{Submanifold} is widely used in 3D perception methods~\cite{PointGroup,MTML,MASC,3D-MPA} to extract features from point clouds. Following the common practice, we first convert the point cloud data into regular volumetric grids. Then a UNet-like structure~\cite{UNet} composed of stacked 3D sparse convolution layers~\cite{Submanifold} is used to extract voxel features $F_\text{voxel}$. Third, we map the voxel features $F_\text{voxel}$ back to point features  $F_\text{point}$.
Based on point features $F_\text{point}$, two branches are built, one for predicting point labels and the other for predicting the per-point center shift vectors.

\paragraph{Semantic Label Prediction Branch}
We apply a 2-layer Multi-Layer Perception (MLP) with a softmax layer upon $F_\text{point}$ to produce semantic scores for every class. The class with the highest score will be regraded as the predicted point label. The cross-entropy loss of semantic scores $\mathcal{L}_\text{seg}$ is used to train this branch.

\paragraph{Center Shift Vector Prediction Branch}
Paralleled with the semantic label prediction branch, we apply a 2-layer MLP upon  $F_\text{point}$ to predict the point-wise center shift vector $\triangle x_i$ ($\triangle x_i \in \mathbb{R}^3$), which represents the offset from each point to its instance center, similar to \cite{3D-MPA,PointGroup}. The instance center is defined as the coordinate mean of all points in this instance.
During training, $\mathcal{L}_\text{shift}$ is used to optimize the center shift vector prediction, which is formulated as

\begin{equation}
\begin{aligned}
 &\mathcal{L}_\text{shift} = \frac{1}{\sum \limits_{p_i \in P} \mathbbm{1}(p_i \in P_\text{fg})} \cdot   \sum  \limits_{p_i \in P} \mathcal{L}(p_i)\text{,} \\
 &\mathcal{L}(p_i) =  w(p_i) \cdot {\parallel \triangle x^\text{gt}_i - \triangle x^\text{pred}_i \parallel}_1 \cdot  \mathbbm{1}(p_i \in P_\text{fg})     \text{,} \\
 & w(p_i) = \min({\parallel\triangle x^\text{gt}_i\parallel}_2, 1) \text{.} \\
\end{aligned}
\label{equ:Lshift}
\end{equation}

$\mathbbm{1}(\cdot)$ is the indicator function. 
$P$ and $P_\text{fg}$ denote the whole point set and the foreground point set respectively. Background points are ignored in $\mathcal{L}_\text{shift}$.
$w(p_i)$ serves as a point-wise weighted term.
Points closer to the instance center less rely on the center shift vectors and 
should contribute less to the loss.
\begin{figure*}[thbp]
    \centering
    
    \subfigure[original coordinates]
    {\begin{minipage}[p]{.23\linewidth}
        \centering
        \includegraphics[width=\linewidth]{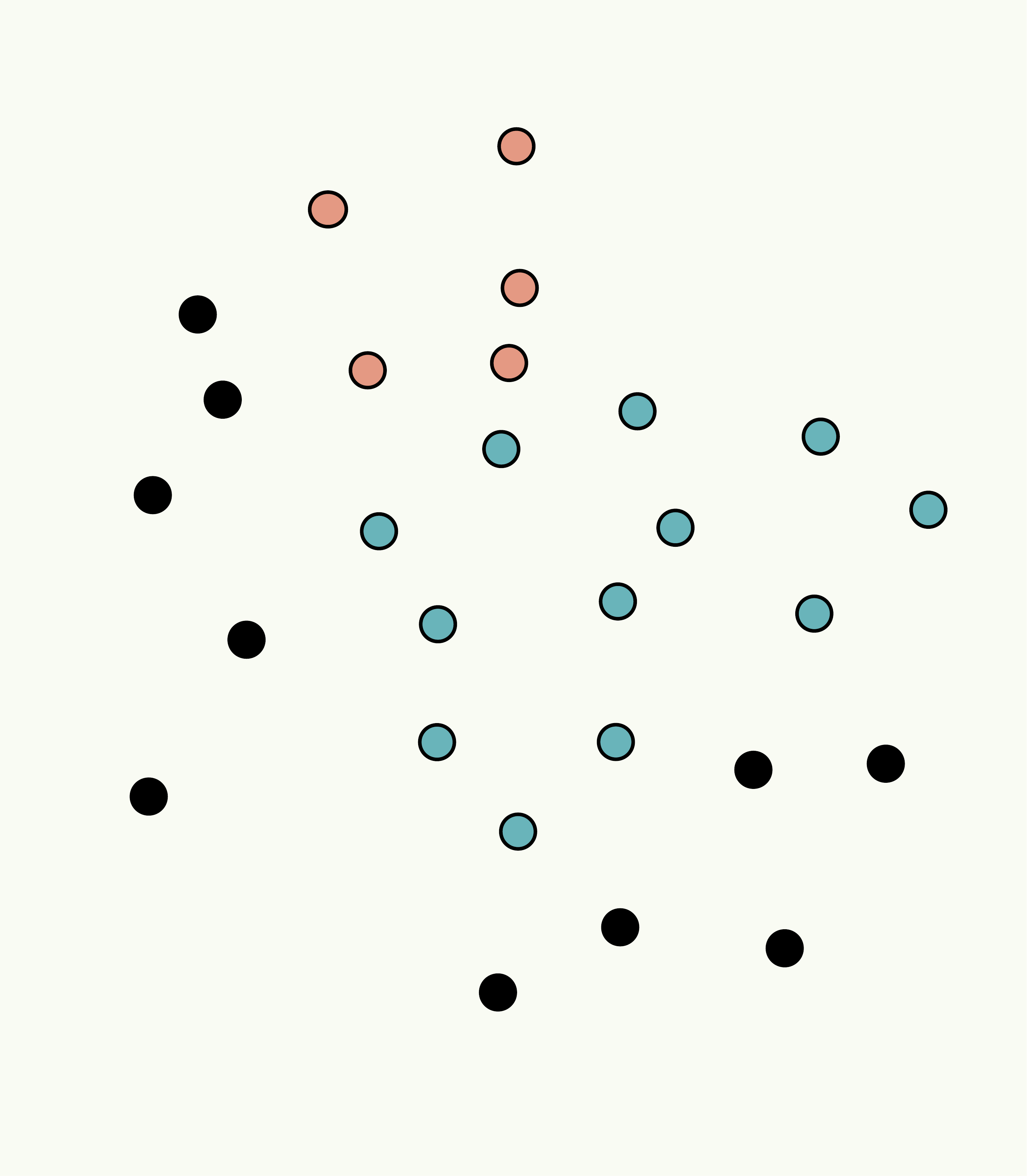}
        \label{fig:a}
    \end{minipage}}
    \subfigure[shifted coordinates]
    {\begin{minipage}[p]{.23\linewidth}
        \centering
        \includegraphics[width=\linewidth]{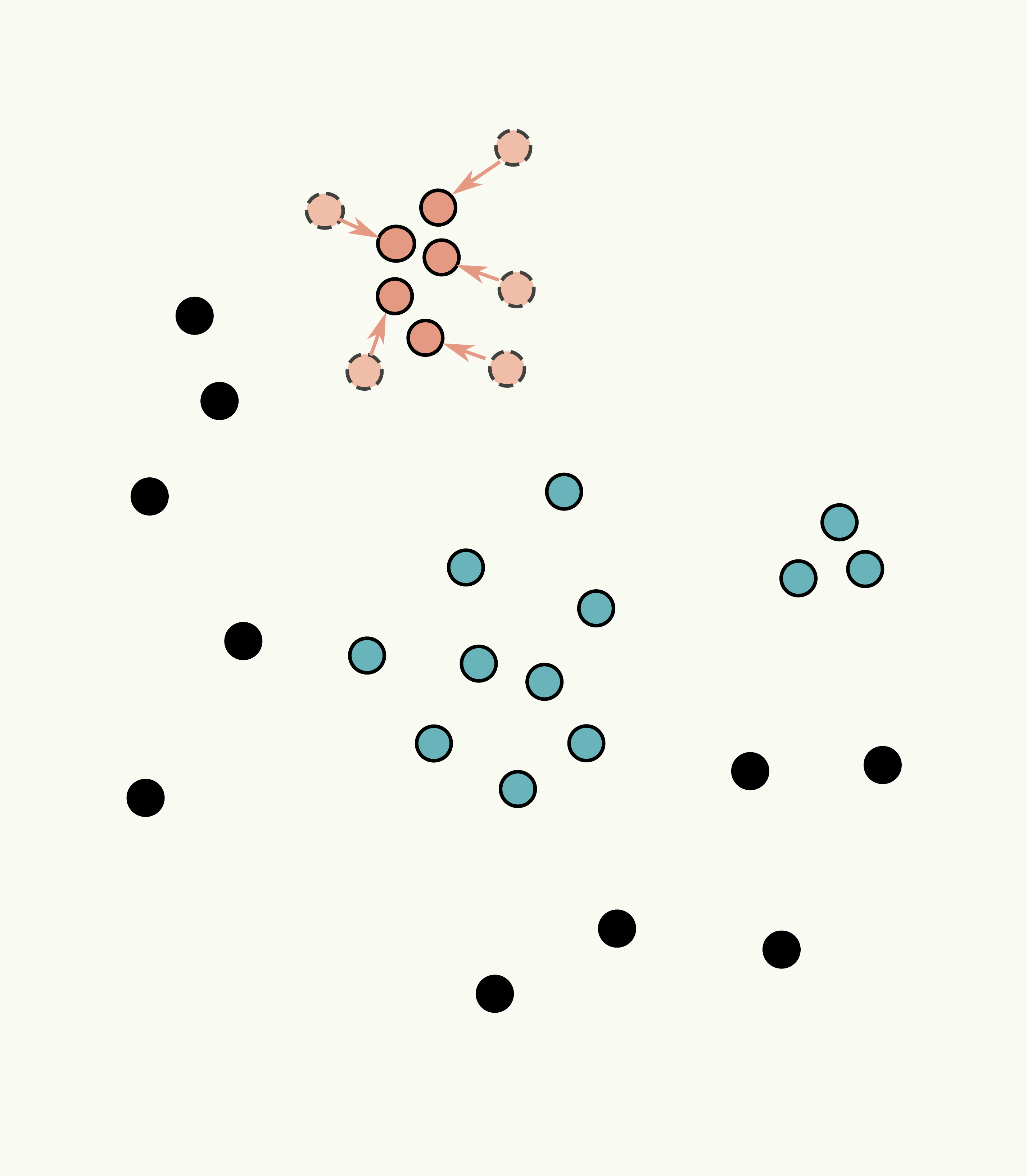}
        \label{fig:b} 
    \end{minipage}} 
    \subfigure[point aggregation]
    {\begin{minipage}[p]{.23\linewidth}
        \centering
        \includegraphics[width=\linewidth]{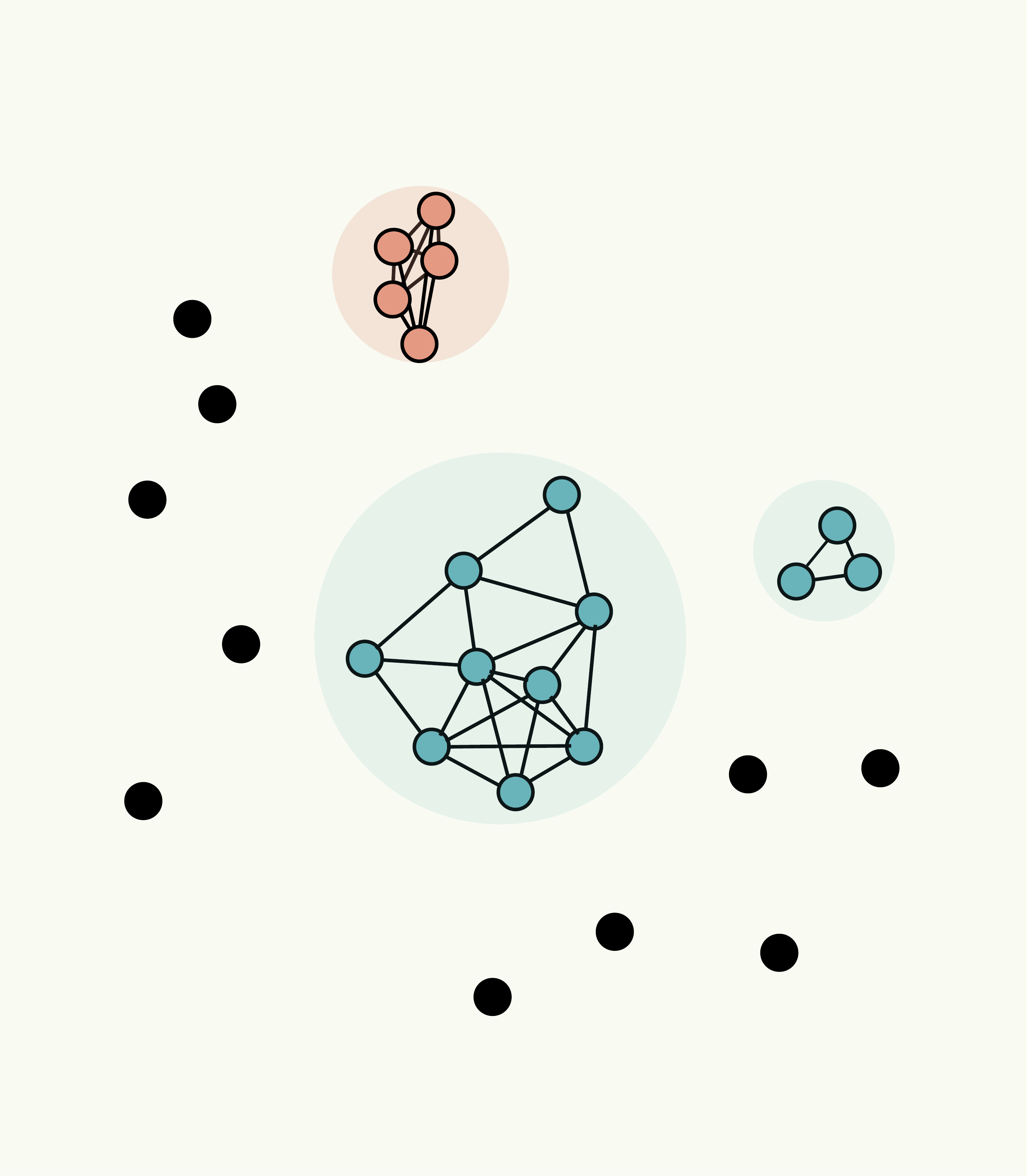}
        \label{fig:c}
    \end{minipage}} 
    \subfigure[set aggregation]
    {\begin{minipage}[p]{.23\linewidth}
        \centering
        \includegraphics[width=\linewidth]{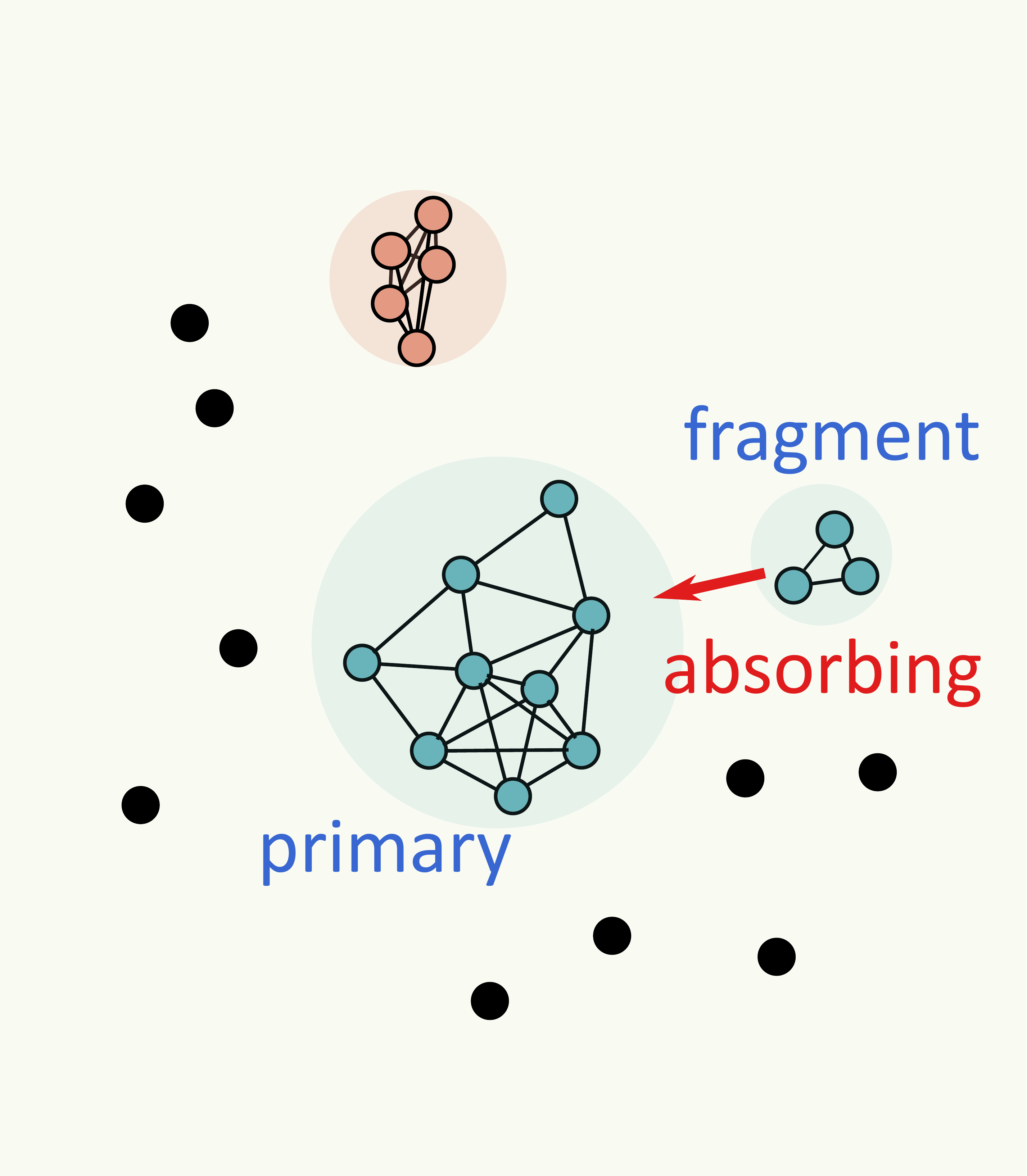}
        \label{fig:d} 
    \end{minipage}} 
    \caption{Illustrations of hierarchical aggregation. Points with different colors belong to different categories. Black points belong to background. (a): Points distributed in real 3D space. (b): After applying the center shift vector to each point, points belong to the same instance are closer in 3D space. (c): Point aggregation. Aggregating  points into sets based on fixed spatial clustering bandwidth. (d): Set aggregation. Primary instances absorb surrounding fragments with dynamic clustering bandwidth to form complete instances. }
\end{figure*}

\subsection{Point Aggregation}
\label{sec:point-aggr}
In the 3D space, points of the same instance are inherently adjacent to each other. It's intuitive to utilize this spatial constraint for clustering.
Thus, based on the semantic label and the center shift vector, we use a basic and compact clustering method to get preliminary instances. First, as shown in Fig.~\ref{fig:b}, according to the point-wise center shift vector $\triangle x_i$, we shift every point $x^\text{origin}_i$ toward its instance center, making the points of the same instance spatially closer to each other. The shifted coordinate is computed as,
\begin{equation}
       {x^\text{shift}_i} = {x^\text{origin}_i} + \triangle x_i\text{.}
\end{equation}

Second, we ignore background points and regard each foreground point as a node. For every pair of nodes, if they have the same semantic label and their spatial distance is smaller than a fixed spatial clustering bandwidth $r_\text{point}$, an edge between these two nodes is created.
After traversing all the pairs of node and establishing edges, the whole point cloud is separated into multiple independent sets, as shown in Fig.~\ref{fig:c}. Each set can be viewed as a preliminary instance prediction.

\subsection{Set Aggregation}
\label{sec:set-aggr}
\begin{figure}[thbp]
    \centering
    \includegraphics[width=\linewidth]{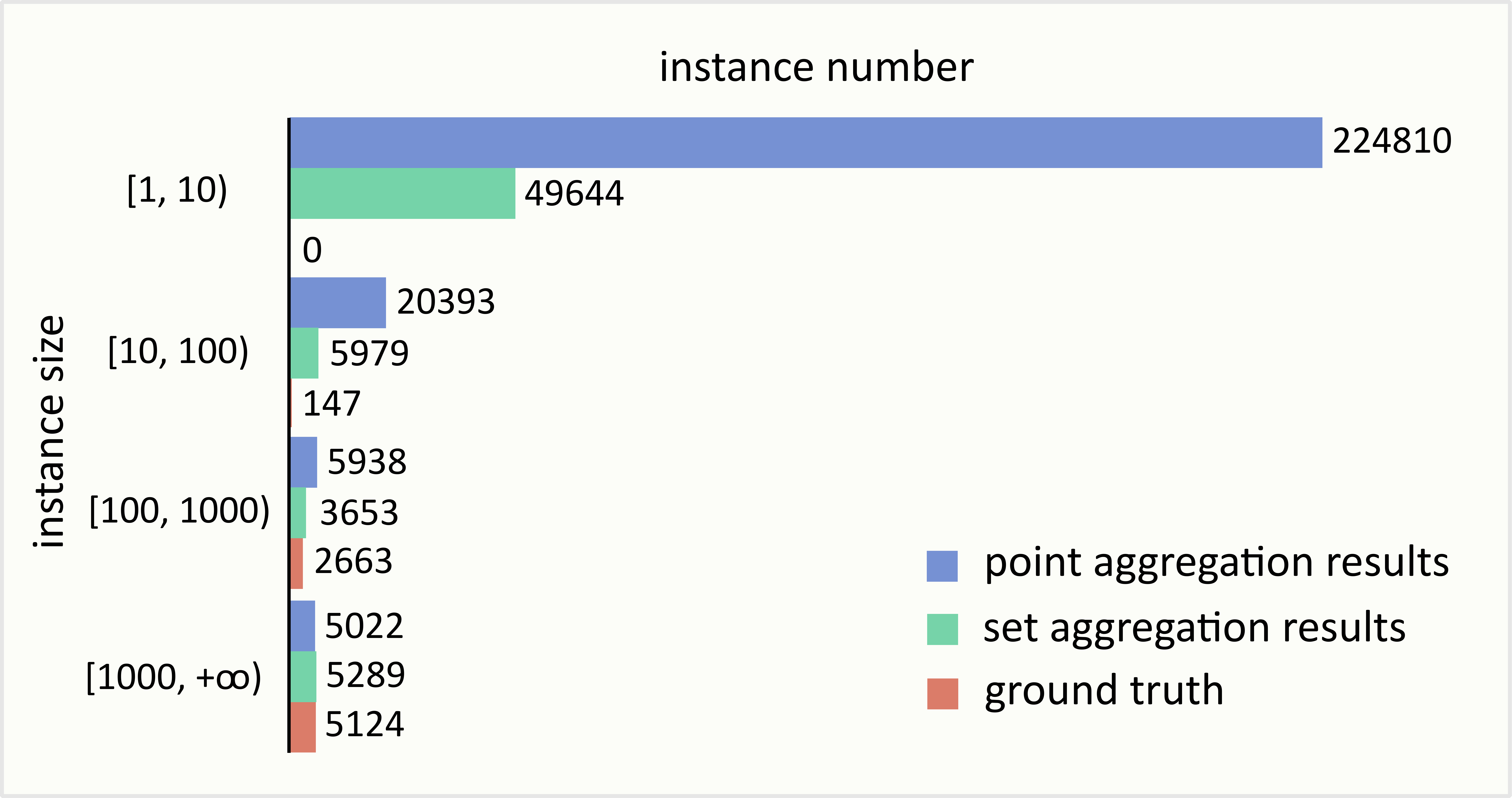}
    \caption{Distribution of instance size. The instance size is defined as the number of points inside an instance. Blue, green and red correspond to point aggregation results, set aggregation results and ground truth, respectively. The statistics are based on the ScanNet v2~\cite{ScanNet} validation set.\label{fig:instance_size}}
\end{figure}

Fig.~\ref{fig:instance_size} shows the distribution of instance size (the number of points in an instance) of the ground truth and the point aggregation results. Compared with the sizes of ground truth instances, point aggregation generates a much larger number of instance predictions with small sizes. 
It is because center shift vectors are not totally accurate. Point aggregation cannot guarantee that all the points in an instance are grouped together. 
% And we adopt a low clustering bandwidth for point aggregation for preventing under-segmentation.
As illustrated in Fig.~\ref{fig:d}, most points with accurate center shift vectors can be clustered together to form incomplete instance predictions. We call these instances ``primary instances''.
But a minority of points with poor center shift vector predictions split from the majority and form fragmentary instances with small sizes, which we call ``fragments''.
Fragments are too small in size to be regarded as complete instances, but are possible to be the missing part of primary instances. Considering the large number of fragments, it's not appropriate to directly filter out fragments with a hard threshold. Intuitively, we can aggregate primary instances and fragments at set level to generate complete instance predictions.

Based on the above clues, we propose the set aggregation to smooth the instance predictions generated by point aggregation, which is shown in Fig.~\ref{fig:d}. The detailed procedure is provided in Alg.~\ref{alg:set_aggr}.
Briefly, if the following two conditions are satisfied, we consider the fragment $m$ to be a part of the primary instance $n$.
First, among all the primary instances that have the same semantic label with the fragment $m$, the primary instance $n$ is the one whose geometric center is closest to the fragment $m$. 
Secondly, for the fragment $m$ and the primary instance $n$, the distance between their geometric center should be smaller than $r_\text{set}$, which is the dynamic clustering bandwidth defined as,
\begin{equation}
\begin{aligned}
    r_\text{set} &= \max(r_\text{size}, r_\text{cls}),\\
    r_\text{size} &= \alpha\sqrt{S^{n}_\text{prim}}.\\
\end{aligned}
\label{equ:thre}
\end{equation}
The clustering bandwidth of set aggregation is determined by $r_\text{size}$ and $r_\text{cls}$.
$r_\text{size}$ denotes the size-specific bandwidth.
It is reasonable that the larger primary instances should absorb fragments in a wider range, and we consider $r_\text{size}$ relative to the square root of the primary instance $n$’s size $S^{n}_\text{prim}$. 
$r_\text{cls}$ denotes the class-specific bandwidth, which is the statistical average instance radii of the specific class.
The distribution of instance size after the set aggregation is shown in Fig.~\ref{fig:instance_size}. A large amount of fragments are combined together with primary instances to form instances with higher quality.

\begin{algorithm}[htb]  
  \caption{Set aggregation. $N_\text{frag}$ is the number of fragments. $N_\text{prim}$ is the number of primary instances.}  
  \label{alg:set_aggr}  
    \KwData{\\
    Fragments: $\{I^1_\text{frag}, I^2_\text{frag},..., I^{N_\text{frag}}_\text{frag}\}$\\
    Primary instances:  $\{I^1_\text{prim}, I^2_\text{prim},..., I^{N_\text{prim}}_\text{prim}\}$\\
    Centers of fragments: $\{c^1_\text{frag}, c^2_\text{frag},..., c^{N_\text{frag}}_\text{frag}\}$\\
    Centers of primary instances: $\{c^1_\text{prim}, c^2_\text{prim},..., c^{N_\text{prim}}_\text{prim}\}$\\
    Class labels of fragments : $\{L^1_\text{frag}, L^2_\text{frag},..., L^{N_\text{frag}}_\text{frag}\}$\\
    Class labels of primary instances: $\{L^1_\text{prim}, L^2_\text{prim},..., L^{N_\text{prim}}_\text{prim}\}$\\
    Dynamic set aggregation bandwidths: $\{r^1_\text{set}, r^2_\text{set},..., r^{N_\text{prim}}_\text{set}\}$\\}
    \KwResult{
    \\
    A set of refined instances: 
    $\{I^1_\text{prim}, I^2_\text{prim},..., I^{N_\text{prim}}_\text{prim}\}$}
    
    \For{$m = 1 \to N_{\rm frag}$}{
         ${\rm index} = -1$\\
         $d_\text{min} =  +\infty$\\
        \For{$n = 1 \to N_{\rm prim}$}
        {
            \If{ {$L^m_{\rm frag} == L^n_{\rm prim}$ {\rm and}\\  $\|c^m_{\rm frag} - c^n_{\rm prim}\| < d_{\rm min}$}  }
            {
                 ${\rm index} = n$\\
                 $d_\text{min} = \|c^m_\text{frag} - c^n_\text{prim}\|$\\
            }
        }
        \If { $ d_{\rm min} < r^{\rm index}_{\rm set}$}  
        {     
            $I^\text{index}_\text{prim} =                I^\text{index}_\text{prim} \cup I^m_\text{frag}$
        }
    }
    \Return{$\{I^1_{\rm prim}, I^2_{\rm prim},..., I^{N_{\rm prim}}_{\rm prim}\}$}
\end{algorithm}

\begin{figure}[htb]
\centering

\includegraphics[width=0.99\linewidth]{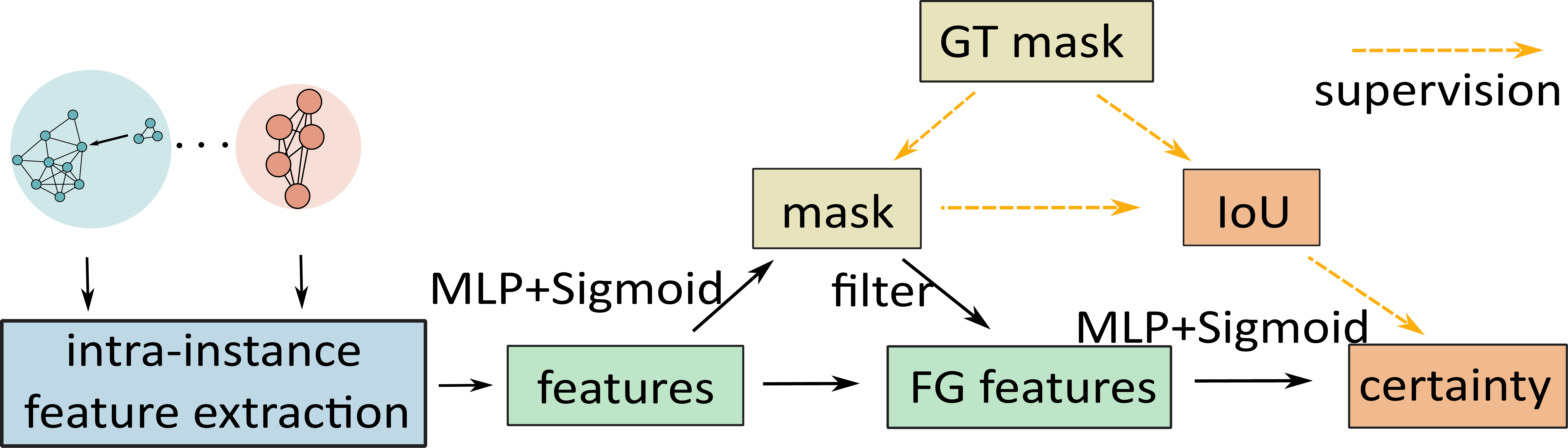}
\caption{The details of the intra-instance prediction network.}
\label{fig:intra}

\end{figure}

\begin{table*}[thbp]
    \centering
    \resizebox{1\linewidth}{!}{
    \setlength{\tabcolsep}{4pt}{
    \begin{tabular}{llllllllllllllllllll}
    \toprule
     Method  & $AP_{50}$ & \rotatebox{90}{bathtub} & \rotatebox{90}{bed} & \rotatebox{90}{bookshe.} & \rotatebox{90}{cabinet} & \rotatebox{90}{chair} & \rotatebox{90}{counter} & \rotatebox{90}{curtain} & \rotatebox{90}{desk} & \rotatebox{90}{door} & \rotatebox{90}{otherfu.} & \rotatebox{90}{picture} & \rotatebox{90}{refrige.} & \rotatebox{90}{s. curtain} & \rotatebox{90}{sink} & \rotatebox{90}{sofa} & \rotatebox{90}{table} & \rotatebox{90}{toilet} & \rotatebox{90}{window} \\
    \midrule
    DPC~\cite{DPC}&35.5&50.0&51.7&46.7&22.8&42.2&13.3&40.5&11.1&20.5&24.1&7.5&23.3&30.6&44.5&43.9&45.7&97.4&23.0\\
    3D-SIS~\cite{3DSIS}&38.2&100.0&43.2&24.5&19.0&57.7&1.3&26.3&3.3&32.0&24.0&7.5&42.2&85.7&11.7&69.9&27.1&88.3&23.5\\
    MASC~\cite{MASC}&44.7&52.8&55.5&38.1&38.2&63.3&0.2&50.9&26.0&36.1&43.2&32.7&45.1&57.1&36.7&63.9&38.6&98.0&27.6\\
    PanopticFusion~\cite{PanopticFusion}&47.8&66.7&71.2&59.5&25.9&55.0&0.0&61.3&17.5&25.0&43.4&43.7&41.1&85.7&48.5&59.1&26.7&94.4&35.0\\
    3D-BoNet~\cite{BoNet}&48.8&100.0&67.2&59.0&30.1&48.4&9.8&62.0&30.6&34.1&25.9&12.5&43.4&79.6&40.2&49.9&51.3&90.9&43.9\\
    MTML~\cite{MTML}&54.9&100.0&80.7&58.8&32.7&64.7&0.4&81.5&18.0&41.8&36.4&18.2&44.5&100.0&44.2&68.8&57.1&100.0&39.6\\
    3D-MPA~\cite{3D-MPA}&61.1&100.0&83.3&76.5&52.6&75.6&13.6&58.8&\textbf{47.0}&43.8&43.2&35.8&65.0&85.7&42.9&76.5&55.7&100.0&43.0\\
    PointGroup~\cite{PointGroup}&63.6&100.0&76.5&62.4&50.5&79.7&11.6&69.6&38.4&44.1&55.9&47.6&59.6&100.0&\textbf{66.6}&75.6&55.6&99.7&51.3\\
    GICN~\cite{GICN}&63.8&100.0&\textbf{89.5}&80.0&48.0&67.6&14.4&73.7&35.4&44.7&40.0&36.5&\textbf{70.0}&100.0&56.9&\textbf{83.6}&59.9&\textbf{100.0}&47.3\\
    \thename & \textbf{69.9}&\textbf{100.0}&84.9&\textbf{82.0}&\textbf{67.5}&\textbf{80.8}&\textbf{27.9}&\textbf{75.7}&46.5&\textbf{51.7}&\textbf{59.6}&\textbf{55.9}&60.0&\textbf{100.0}&65.4&76.7&\textbf{67.6}&99.4&\textbf{56.0}\\
    \bottomrule
    \end{tabular}}
    }
    \caption{Quantitative comparison on the testing set of ScanNet v2~\cite{ScanNet} benchmark. Our \thename\ achieves the SOTA performance, outperforming all other methods by a large margin.}
    \label{tab:main_result_scannet}
\end{table*}

\begin{table}[]
    \centering
    \begin{tabular}{lcccc}
    \toprule
     Method  & mCov &  mWCov & mPrec & mRec\\
    \midrule
    SGPN$^{\dagger}$~\cite{SGPN}& 32.7 & 35.5 & 36.0 & 28.7\\
    ASIS$^{\dagger}$~\cite{ASIS}& 44.6 & 47.8 & 55.3 & 42.4\\
    PointGroup$^{\dagger}$~\cite{PointGroup} & - & - & 61.9 & 62.1\\
    \thename$^{\dagger}$ & \textbf{64.3} & \textbf{66.0} & \textbf{71.1} & \textbf{65.0} \\
    \toprule
    SGPN$^{\ddagger}$~\cite{SGPN}& 37.9 & 40.8  & 38.2 & 31.2\\
    PartNet$^{\ddagger}$~\cite{PartNet} & - & - & 56.4 & 43.4\\
    ASIS$^{\ddagger}$~\cite{ASIS}& 51.2 & 55.1 & 63.6 & 47.5\\
    3D-BoNet$^{\ddagger}$~\cite{BoNet}& - & - & 65.6 & 47.6\\
    OccuSeg$^{\ddagger}$~\cite{OccuSeg}& - & - & 72.8 & 60.3\\
    GICN$^{\ddagger}$~\cite{GICN} & - & - & 68.5 & 50.8\\
    PointGroup$^{\ddagger}$~\cite{PointGroup} & - & - & 69.6 & 69.2\\
    \thename$^{\ddagger}$ & \textbf{67.0} & \textbf{70.4} & \textbf{73.2} & \textbf{69.4} \\
    \bottomrule
    \end{tabular}
    \caption{Quantitative comparison on S3DIS~\cite{S3DIS}. Methods marked with $\dagger$ are evaluated on Area 5 and those marked with $\ddagger$ are on the 6-fold cross validation. Our method significantly outperforms previous methods in terms of mCov (coverage), mWCov (weighted coverage), mean precision (mPrec) and mean recall (mRec).}
    \label{tab:main_result_S3DIS}
\end{table}

% \begin{table}[]
%     \centering
%     % \resizebox{1\linewidth}{!}{
%     % \setlength{\tabcolsep}{4pt}{
%     \begin{tabular}{lcccc}
%     \toprule
%     Method  & mCov &  mWCov & mPrec & mRec\\
%     \midrule
%     SGPN~\cite{SGPN}& 37.9 & 40.8  & 38.2 & 31.2\\
%     PartNet~\cite{PartNet} & - & - & 56.4 & 43.4\\
%     ASIS~\cite{ASIS}& 51.2 & 55.1 & 63.6 & 47.5\\
%     3D-BoNet~\cite{BoNet}& - & - & 65.6 & 47.6\\
%     OccuSeg~\cite{OccuSeg}& - & - & 72.8 & 60.3\\
%     GICN~\cite{GICN} & - & - & 68.5 & 50.8\\
%     PointGroup~\cite{PointGroup} & - & - & 69.6 & 69.2\\
%     \thename & \textbf{67.0} & \textbf{70.4} & \textbf{73.2} & \textbf{69.4} \\
%     \bottomrule
%     \end{tabular}
%     % }
%     \caption{Quantitative comparison on S3DIS~\cite{S3DIS}. Methods marked with $\dagger$ are evaluated on Area 5 and those marked with $\ddagger$ are on the 6-fold cross validation. Our method significantly outperforms previous methods in terms of mCov (coverage), mWCov (weighted coverage), mean precision (mPrec) and mean recall (mRec).}
%     \label{tab:main_result_S3DIS}
% \end{table}

\subsection{Intra-instance Prediction Network}
\label{sec:intra}
The hierarchical aggregation may mistakenly absorb fragments belonging to other instances, generating inaccurate instance predictions.
Thus, we propose the intra-instance prediction network for further refining the instances, as shown in Fig.~\ref{fig:intra}. First, we crop instance point cloud patches as input and use the 3D submanifold sparse convolution network to extract features inside instances. 
After intra-instance feature extraction, the mask branch predicts binary masks to distinguish the instance foreground and background.
For every predicted instance, we choose the best matched GT (ground truth) as the mask supervision. The overlapped parts between the predicted instance and the GT are assigned with positive labels, and others are assigned with negative labels. 
Low quality instances (low IoU with GT) contain little instance-level information and are valueless to optimizing the mask branch. 
Thus, only the instances with IoU higher than $0.5$ are used as training samples, while others are ignored. 
For mask prediction, the loss is formulated as,
\begin{equation}
\begin{aligned}
    \mathcal{L}_\text{mask} = & - \frac{1}{\sum\limits^{N_\text{ins}}\limits_{i=1} \mathbbm{1}(iou_i > 0.5) \cdot N_{i}}  \cdot  \sum\limits^{N_\text{ins}}\limits_{i=1}\Big \{ \mathbbm{1}(iou_i > 0.5) \\
     & \cdot \sum\limits^{N_i}\limits_{j=1} \big [ y_j \cdot \log(\hat{y}_j) + (1-y_j)\cdot \log(1- \hat{y}_j) \big ] \Big \},
\end{aligned}
\label{equ:Lmask}
\end{equation}
where $N_\text{ins}$ denotes the number of instances and $N_{i}$ denotes the point number of instance $i$.

Besides mask prediction, the instance certainty score is needed for ranking among instances. We utilize masks for better scoring instances, as illustrated in Fig.~\ref{fig:intra}.
Firstly, masks are used to filter out features of background point, which would be noise for scoring. The remained foreground features are sent into a MLP with a sigmoid layer to predict the instance certainty scores. Secondly, inspired by \cite{MSRCNN, IOUNet, GS3D}, we regard the IoUs between predicted masks and GT masks as the mask quality and use them to supervise instance certainty.
For the score prediction, the loss is formulated as,
\begin{equation}
\begin{aligned}
     \mathcal{L}_\text{score} = & - \frac{1}{N_\text{ins}} \cdot
     \sum\limits^{N_\text{ins}}\limits_{i=1} \big [ iou_i \cdot \log(\hat{s}_i) \\
     & + (1-iou_i)\cdot \log(1- \hat{s}_i) \big ].
\end{aligned}
\label{equ:Lscore}
\end{equation}
Ablation studies in Sec.~\ref{sec:ablation} further demonstrate that using masks to assist score prediction boosts performance.

\subsection{Multi-task Training}
The whole network is trained from scratch in an end-to-end manner and optimized by a joint loss consisting of several loss terms,
\begin{equation}
     \mathcal{L} =  \mathcal{L}_\text{seg} + \mathcal{L}_\text{shift} + \mathcal{L}_\text{mask} +  \mathcal{L}_\text{score},
\end{equation}
where $\mathcal{L}_\text{seg}$ is the cross-entropy loss of semantic scores, and $\mathcal{L}_\text{shift}$, $\mathcal{L}_\text{mask}$ and $\mathcal{L}_\text{score}$ are defined in Eq.~\ref{equ:Lshift}, \ref{equ:Lmask} and \ref{equ:Lscore} respectively.

\subsection{NMS-free and Single-forward Inference}
Proposal-based methods~\cite{3D-MPA,GICN} usually require dense proposals for better covering instances. And many clustering-based methods~\cite{MTML,PointGroup} adopt multiple clustering strategies to generate redundant instance predictions. Thus, non-maximum suppression (NMS) or other post-processing steps which function as NMS, are widely required for removing duplicated instance predictions.
But in our \thename, one point is only clustered into a single instance in point aggregation, resulting in no overlap among instance predictions. We can directly use instance certainty scores to rank the instances and take the ones with the highest scores as final predictions, not requiring any post-processing steps.
Besides, iterative clustering procedure is widely used in clustering based methods~\cite{OccuSeg, JSIS3D}, which refines predictions step by step but is time-consuming. In contrast, \thename\ only requires a compact single-forward inference procedure to generate accurate predictions.
With the NMS-free and single-forward design, \thename\ keeps a much more concise pipeline with higher efficiency. 

\section{Experiments}
In this section, we first present our experimental settings (Sec.~\ref{sec:exp_setting}).
Then, we provide both quantitative (Sec.~\ref{sec:quant}) and qualitative evaluations (Sec.~\ref{sec:quali}) to demonstrate the effectiveness of \thename.
To better validate each component of our method, we provide detailed ablation studies (Sec.~\ref{sec:ablation}).
And evaluation on inference speed (Sec.~\ref{sec:inference_speed}) is offered to prove the efficiency of \thename.

\subsection{Experimental Settings}
\label{sec:exp_setting}
\paragraph{ScanNet v2} 
The ScanNet v2~\cite{ScanNet} dataset 
is the most accepted and robust dataset in 3D instance segmentation.
It contains $1,613$ scans with 3D object instance annotations. The dataset is split into the training, validation and testing set, each with $1,201$, $312$, and $100$ scans, respectively. $18$ object categories are used for instance segmentation evaluation. 
To fairly compare with other works, we report results on the testing set which come from the official evaluation server.
For ablation studies, we report results on the validation set. 
Keeping the same with the ScanNet v2 benchmark, we use the mean average precision with an IoU threshold of $0.5$ ($AP_{50}$) as the main evaluation metric.
We also report the mean average precision at the overlap $0.25$ ($AP_{25}$) and overlaps from $0.5$ to $0.95$ ($AP$) in the ablation study.

\paragraph{S3DIS} 
To validate the generalization of \thename, we also conduct experiments on
the S3DIS~\cite{S3DIS} dataset. S3DIS has 3D scans across six areas with 271 scenes in total. 
Each point is assigned with one label out of 13 semantic classes. All the 13 classes are used in instance evaluation.
We report results evaluated on both Area 5 and the 6-fold cross  validation.
We use coverage (mCov), weighted coverage (mWCov), mean precision (mPrec) and mean recall (mRec) with the IoU threshold of 0.5 as evaluation metrics.

\paragraph{Experimental details} 
Our model is trained on one single Titan X GPU with a batch size of $4$ for $120k$ iterations. The initial learning rate is $0.001$ and decays with a cosine anneal schedule~\cite{CosineLr}. For stability and efficiency, we do not adopt set aggregation during the training phase. And this does not affect the effectiveness of set aggregation during inference.
We set the voxel size as $0.02$m following the common practice \cite{MASC, PointGroup}. The bandwidth $r_\text{point}$ for point aggregation is set as $0.03$.
$\alpha$ in Eq.~\ref{equ:thre} is set as $0.01$.
The final predictions containing less than $100$ points are filtered out before evaluation.

\begin{figure*}[htb]
\centering
\includegraphics[width=\linewidth]{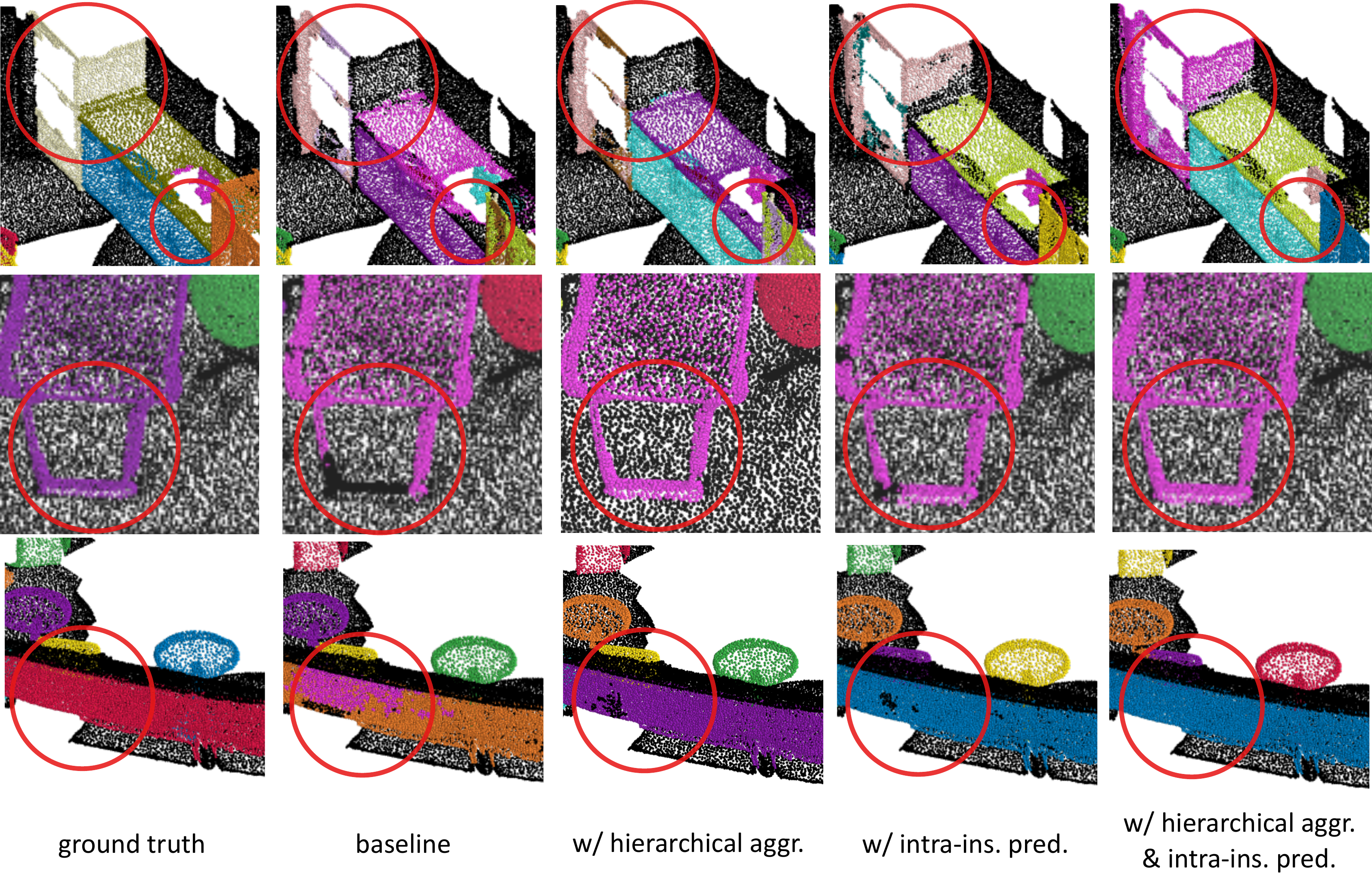}
\caption{Qualitative results of challenging cases of ScanNet v2~\cite{ScanNet}. Key regions are circled with red.
The hierarchical aggregation and intra-instance prediction contribute to more fine-grained predictions, especially for objects with large sizes and fragmentary point clouds.}
\label{fig:visualization_detail}
\end{figure*}

\subsection{Quantitative Evaluation}
\label{sec:quant}
\paragraph{ScanNet v2}
In  Tab.~\ref{tab:main_result_scannet}, we compare our \thename\ with other methods on the unreleased testing set of ScanNet v2~\cite{ScanNet} benchmark. \thename\ achieves the highest $AP_{50}$ of $69.9\%$, ranking the first place on the leaderboard of ScanNet v2 and surpassing the previous state-of-the-art (SOTA) work~\cite{GICN} by $6.1\%$.
For results on each class, our method achieves the best performance in $12$ out of $18$ classes.

\paragraph{S3DIS} 
In Tab.~\ref{tab:main_result_S3DIS}, we present the results on S3DIS. \thename\ achieves much higher results than other methods in terms of all the widely-used metrics (mCov, mWCov, mPre and mRec).
ScanNet v2 and S3DIS are quite different in terms of category, scene style and point cloud density. The SOTA performances of \thename\ on both ScanNet v2 and S3DIS prove the high generalization ability.

\subsection{Qualitative Evaluation}
\label{sec:quali}
Fig.~\ref{fig:visualization_detail} visually shows the effectiveness of the hierarchical aggregation and intra-instance prediction.
For objects with large sizes and fragmentary point clouds, grouping all points together is quite challenging.
We can observe that with the proposed hierarchical aggregation and intra-instance prediction,  precise instance segmentation masks are obtained.

\subsection{Ablation Study}
\label{sec:ablation}
To validate the design of \thename, we perform a series of ablation studies on the ScanNet validation set.
\paragraph{Ablation on the Hierarchical Aggregation and Intra-instance Prediction Network}
Tab.~\ref{tab:ab_inter_intra} proves the effectiveness of the hierarchical aggregation and intra-instance prediction network. The intra-instance prediction promotes results by $2.7\% \ AP$, $2.4\% \ AP_{50}$ and $0.3\% \ AP_{25}$ . And the set aggregation further improves results by $1.0\% \ AP$, $0.7\% \ AP_{50}$ and $0.7\% \ AP_{25}$.

\begin{table}[thbp]

    \centering\small 
    \setlength{\tabcolsep}{2pt}{
    \begin{tabular}{cccccc}
    \toprule
    Point aggr. & Set aggr. & Intra-ins. pred.  & $AP$ & $AP_{50}$ & $AP_{25}$\\ 
    \midrule
    \checkmark & & & 39.8 & 61.0 & 74.6 \\
    %  & 40.1 & 61.3 & 74.7 \\
    \checkmark & &\checkmark &  42.5 & 63.4 & 74.9\\
    \checkmark & \checkmark & \checkmark & 43.5 & 64.1 & 75.6\\
    \bottomrule
    \end{tabular}}
    \caption{Ablation results on the ScanNet v2 validation set. The hierarchical aggregation and intra-instance prediction bring significant gains in terms of $AP$, $AP_{50}$ and $AP_{25}$.}
    \label{tab:ab_inter_intra}
\end{table}

\begin{table}[thbp]
    \centering\small 
    \begin{tabular}{cccc}
    \toprule
     \makecell{ Using masks to filter\\ points and calculate IoU}  & $AP$ & $AP_{50}$ & $AP_{25}$\\ 
    \midrule
     & 41.9 & 63.1 & 74.9\\
    \checkmark & 43.5 & 64.1 & 75.6\\
    \bottomrule
    \end{tabular}
    \caption{Ablation results on ScanNet v2 validation set for evaluating the effectiveness of using masks to filter points and calculate IoU.}
    \label{tab:ab_filtering}
\end{table}

\begin{table}[thbp]
    \centering\small 
    \begin{tabular}{cccc}
    \toprule
    Mask training samples  & $AP$ & $AP_{50}$ & $AP_{25}$\\ 
    \midrule
    All  & 42.8 & 63.3 & 74.4\\
    IoU \textgreater\ 0.5 & 43.5 & 64.1 & 75.6\\
    \bottomrule
    \end{tabular}
    \caption{Ablation results on ScanNet v2 validation set for evaluating the effectiveness of filtering mask training samples.}
    \label{tab:ab_mask_thre}
\end{table}

\begin{table}[thbp]
    \centering\small
    \resizebox{\linewidth}{!}{
    \setlength{\tabcolsep}{2pt}{

    \begin{tabular}{lcc}
    \toprule
     Method  &\makecell{Whole val set\\inference time (sec)} & \makecell{Per-frame\\inference time (msec)}\\ 
    \midrule
    SGPN~\cite{SGPN}&49433&158439\\
    ASIS~\cite{ASIS}&56757&181913\\
    GSPN~\cite{GSPN}& 3963 &12702\\
    3D-SIS~\cite{3DSIS}&38841&124490\\
    3D-BoNet~\cite{BoNet}& 2871&9202\\
    OccuSeg~\cite{OccuSeg}&594&1904\\
    PointGroup~\cite{PointGroup}& 141&452\\
    GICN~\cite{GICN} &  2688 & 8615\\
    \thename& \textbf{128}&\textbf{410}\\
    \bottomrule
    \end{tabular}}}
    \caption{The inference time on the validation set of ScanNet v2. For fair comparison, the inference time is measured on the same type of GPU (Titan X). Our \thename\ achieves much better inference speed than other methods. }
    \label{tab:infer}

\end{table}

\paragraph{Using Masks to Filter Points and Calculate IoU}
In the intra-instance prediction network, masks are used to assist certainty score predictions, \ie{}, filtering out features of background and calculating IoU which is the supervision signal of the instance certainty, as shown in Fig.~\ref{fig:intra}.
An alternative method is directly using the whole instance features without filtering background points to predict scores and using the IoU between the original input instance and the GT to supervise the instance certainty.
Ablation experiments in Tab.~\ref{tab:ab_filtering} show that, using masks to filter points and calculate IoU improves the results.
Filtering features with the mask can avoid the influence of the background noise.
And masks are much more accurate than original input instances. The IoU between the mask and GT is more suitable to be the supervision signal of the certainty score.

\paragraph{Filtering Mask Training Samples}
As shown in Tab.~\ref{tab:ab_mask_thre}, compared with using all the instances as mask training samples, it's better to filter out low quality instances with the IoU threshold of $0.5$.
Low quality instances usually covers few foreground points but a large amount of background points. These instances may bring ambiguity to the instance-level refinement. It's beneficial to filter them out during training.

\subsection{Inference Speed}
\label{sec:inference_speed}
For real-life applications, \eg, mixed reality and autonomous driving, the inference speed of the whole network is of critical importance. 
We evaluate the efficiency of \thename\ and compare it with other methods, as shown in Tab.~\ref{tab:infer}. 
The single scene inference time is highly correlated to the number of points in the point cloud and varies a lot. Following the evaluation method of \cite{BoNet,OccuSeg,GICN}, 
we use the whole validation set inference time of ScanNet v2 for fair comparison on the efficiency. 
\thename\ only takes $128$ seconds to infer all the 312 scans in the validation set, achieving the highest efficiency among all methods. 
On average, per scan inference latency of \thename\ is $410$ ms.
Point-wise prediction network, point aggregation, set aggregation and the intra-instance prediction network takes $172$, $125$, $4$, $109$ ms, respectively.

\section{Conclusion}
We propose \thename, a concise bottom-up approach for 3D instance segmentation.
We introduce the hierarchical aggregation to generate instance predictions in a two-step manner and the intra-instance prediction for more fine-grained instance predictions.
Experiments on ScanNet v2 and S3DIS demonstrate the effectiveness and generalization of our method. \thename\ also retains much better inference speed than all existing methods, showing its practicability in most scenarios, especially latency-sensitive ones.

\noindent \textbf{Acknowledgement}: This work was in part supported by NSFC (No. 61876212 and No. 61733007) and the Zhejiang Laboratory under Grant 2019NB0AB02.

{\small
\bibliographystyle{ieee_fullname}
\bibliography{hais}
}

\appendix

In this supplementary material, we provide the detailed inference latency of main individual components of \thename\ and compare it with other methods. Besides, we provide more qualitative results to demonstrate the effectiveness of \thename.

\section{Detailed Inference Time}
Tab.~\ref{tab:infer_detail} shows the inference time of main components of different methods. Compared with other methods which require time-consuming clustering and post processing procedures,
our \thename\ keeps a much more efficient pipeline. The point-wise prediction network, point aggregation, set aggregation and intra-instance prediction network takes 172, 125, 4 and 109 ms, respectively.

\begin{table*}[htb]
    \centering 
    \caption{Inference time of main components of different methods on the ScanNet v2 validation set. For fair comparison, data in this table is measured on the same type of GPU (Titan X).}
    \small
    \begin{tabular}{cccc}
    \toprule
     Method  & \makecell{Component \\inference time (msec)} & \makecell{Per frame \\inference time (msec)}\\ 
    \midrule
    SGPN~\cite{SGPN}&\makecell{backbone (GPU): 2080\\group merging (CPU): 149000\\ block merging (CPU): 7119}& 158439\\
    \hline
    ASIS~\cite{ASIS}&\makecell{backbone (GPU): 2083\\mean shift (CPU): 172711\\block merging (CPU): 7119}&181913\\
    \hline
    GSPN~\cite{GSPN}&\makecell{backbone (GPU): 1612\\point sampling (GPU): 9559\\neighbour search (CPU): 1500} & 12702 \\
    \hline
    3D-BoNet~\cite{BoNet}&\makecell{backbone (GPU): 2083\\SCN (GPU): 667\\block merging (CPU): 7119} & 9202\\
    \hline
    OccuSeg~\cite{OccuSeg}&\makecell{backbone GPU): 189\\supervoxel (CPU): 1202\\clustering (GPU+CPU): 513}&1904\\
    \hline
    PointGroup~\cite{PointGroup}&\makecell{backbone (GPU): 128\\clustering (GPU+CPU):221 \\ScoreNet (GPU): 103 }  & 452\\
    \hline
    GICN~\cite{GICN} & \makecell{backbone (GPU): 1497\\SCN (GPU): 667\\block merging(CPU): 7119} & 8615\\
    \hline
    \thename&\makecell{point-wise prediction(GPU): 172\\point aggregation(GPU+CPU): 125 \\set aggregation(GPU): 4\\intra-instance prediction(GPU): 109}& 410\\
    \bottomrule
    \label{tab:infer_detail}
    \end{tabular}
\end{table*}

\section{Additional Qualitative Results}
\begin{figure*}[htb]
\centering
\includegraphics[width=\linewidth]{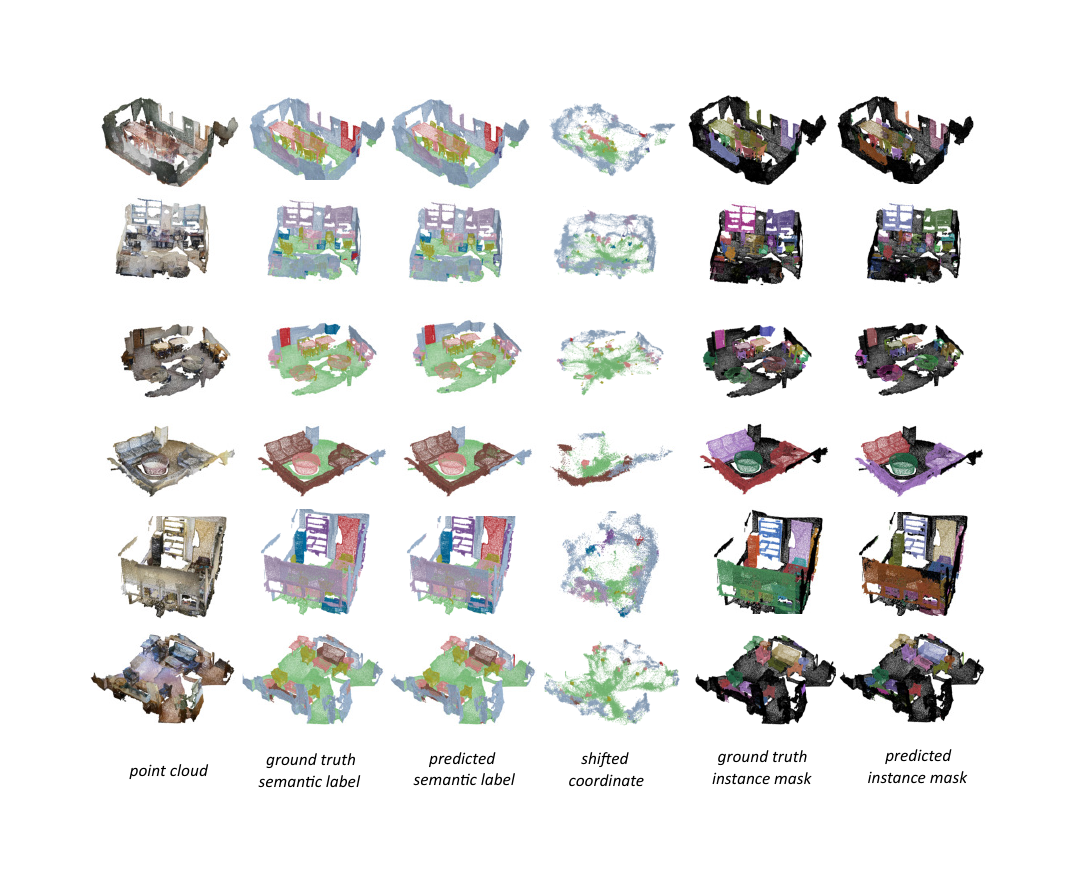}
\caption{Qualitative results on ScanNet v2~\cite{ScanNet}. From left to right: input point cloud, ground truth semantic label, predicted semantic label, shifted coordinate, ground truth instance mask and predicted instance mask. Best viewed in color.}
\label{fig:add_visualization}
\end{figure*}

We show more qualitative results on the validation split of the ScanNet v2 dataset in Fig.~\ref{fig:add_visualization}.
The predicted center shift vectors of some points are not accurate and a large amount of instance fragments come into being. By introducing the hierarchical aggregation and intra-instance prediction, \thename\ generates fine-grained instance predictions.

\end{document}